\documentclass{article}

\usepackage[preprint]{neurips_2026}

\usepackage[utf8]{inputenc} 
\usepackage[T1]{fontenc}    
\usepackage{hyperref}       
\usepackage{url}            
\usepackage{booktabs}       
\usepackage{amsfonts}       
\usepackage{nicefrac}       
\usepackage{microtype}      
\usepackage{xcolor}         
\usepackage{amsmath}
\usepackage{grffile}
\usepackage{subcaption}
\usepackage{appendix}
\usepackage{changes}

\title{Generative Criticality in Large Language Model Temperature Scaling}

\author{%
  Huajian Ruan$^{1,2,3}$ \quad
  Jinyang Li$^{4,5,6}$ \quad
  Xingyu Guo$^{1,2,3,*}$ \quad
  Lingxiao Wang$^{5,7,*}$ \\ [0.4em]
  $^{1}$State Key Laboratory of Nuclear Physics and Technology, Institute of Quantum Matter,\\
  South China Normal University, Guangzhou 510006, China\\
  $^{2}$Key Laboratory of Atomic and Subatomic Structure and Quantum Control (MOE),\\
  Guangdong-Hong Kong Joint Laboratory of Quantum Matter, Guangzhou 510006, China\\
  $^{3}$Guangdong Basic Research Center of Excellence for Structure and Fundamental Interactions of Matter,\\
  Guangdong Provincial Key Laboratory of Nuclear Science, Guangzhou 510006, China\\[0.3em]
  $^{4}$KEK Theory Center, Institute of Particle and Nuclear Studies, Tsukuba, Japan \\
  $^{5}$RIKEN Center for Interdisciplinary Theoretical and Mathematical Sciences (iTHEMS), \\ Wako, Saitama 351-0198, Japan \\
  $^{6}$Graduate University for Advanced Studies (SOKENDAI),\\ Oho 1-1, Tsukuba, Ibaraki 305-0801, Japan \\
  $^{7}$Institute for Physics of Intelligence, The University of Tokyo, \\Hongo,  Tokyo 113-0033, Japan \\
  \texttt{guoxy@m.scnu.edu.cn, lingxiao.wang@riken.jp} \\
  \small{*Corresponding authors}
}

\begin{document}

\maketitle

\begin{abstract}
We propose a statistical-field framework for text generated by large language models (LLMs), treating token embeddings as continuous spin variables on a one-dimensional chain. Defining a susceptibility from the connected two-point correlator and an order parameter from the ensemble-averaged embedding field, we vary the \texttt{softmax} temperature $T$ and observe a sharp susceptibility peak near a characteristic $T_c$ with power-law-like scaling, a concurrent rapid change in the order parameter, and a collapse onto a single semantic direction below $T_c$. The intrinsic dimension estimated by the two nearest neighbor (TwoNN) method independently corroborates these findings, reaching a minimum near $T_c$. Results are robust across model scales (Qwen3: 0.6B--32B) and prompt categories. While the phenomenology closely resembles a continuous phase transition, the non-equilibrium nature of autoregressive generation warrants further investigation. Our framework provides quantitative tools for probing the collective statistical structure of LLM outputs and suggests connections between decoding strategies and critical phenomena.
\end{abstract}

\section{Introduction}
\label{sec:intro}

Understanding the structure of language remains a central challenge in linguistics~\cite{pinker2003language,chomsky2002syntactic} and artificial intelligence. While large language models (LLMs) have enabled text modeling at unprecedented scale~\cite{mitchell2023debate}, most analyses rely on information-theoretic quantities such as entropy and mutual information~\cite{Scheibner2025LargeLM,takahira2016entropy,li1989mutual}, which do not capture microstructural or emergent macroscopic behavior. Recent work has begun mapping LLM-generated texts to statistical field theories~\cite{nakaishi2024critical,sun2025phase}, but rigorous definitions of the relevant physical quantities are still lacking.

We address this gap by constructing a statistical-mechanics framework for LLM-generated text~\cite{carleo2019machine}. Within the token embedding space, we define susceptibility and an order parameter~\cite{amit2005field}, and study their behavior as a function of the \texttt{softmax} temperature $T$~\cite{vaswani2017attention,zhu2024hot}. We find that these quantities exhibit critical behavior~\cite{lin2017critical,chang2023simple}: near a critical temperature $T_c$, the susceptibility diverges and the order parameter undergoes a rapid change. To corroborate this from a geometric perspective, we apply the two nearest neighbor (TwoNN) intrinsic-dimension estimator~\cite{Mendes_Santos_2021,li2018measuring}, which independently identifies the same critical region through non-monotonic features in $I_d(T)$.

Our main contributions are: (i) a statistical-field framework for LLM outputs with well-defined physical observables; (ii) evidence of critical behavior driven by the temperature parameter; and (iii) independent geometric validation via the TwoNN-estimated intrinsic dimension.

\section{LLM Building Blocks and the O(\textit{N}) Model}

An LLM tokenizes input text into subword units via a fixed vocabulary~\cite{bishop2023deep,yang2023teal,mehta2023semantic}, then maps each token to an $N$-dimensional vector through a trainable embedding matrix~\cite{yang2023teal,egger2022text}. Since token embeddings are high-dimensional vectors whose norms concentrate around a characteristic scale, this motivates a heuristic physical picture, treating the text sequence as a one-dimensional lattice chain of $N$-dimensional vectors~\cite{coleman1974spontaneous,amit2005field,eynard1995exact}, for constructing an effective Hamiltonian over the high-dimensional semantic space,

\begin{equation}
\mathcal{H} = \sum_{\sigma,\tau} J_{\sigma\tau}\, t_\sigma\, t_\tau + \sum_\sigma H_\sigma\, t_\sigma\,,
\end{equation}
where $J$ is the coupling matrix (encoding non-local interactions between all pairs of lattice sites), $H$ is an external field set by the prompt, and Greek indices label all different lattice sites. To simplify the problem, we omit the higher-order interaction terms, e.g., $O(t^2)$, but LLMs can still determine the effective couplings and thus governs how each site responds to $H$.

Inter-token interactions are mediated by the self-attention mechanism~\cite{vaswani2017attention}. Given query, key, and value projections $Q_i{=}F_Q(t_i)$, $K_i{=}F_K(t_i)$, $\mathcal{V}_i{=}F_{\mathcal{V}}(t_i)$:
\begin{equation}
\mathrm{Attention}(Q,K,\mathcal{V}) = \mathrm{softmax}\!\bigl(QK^T/\sqrt{d_k}\bigr)\,\mathcal{V}\,,
\end{equation}
where $d_k$ is the dimension of the key vectors. Stacking $M$ layers of $\mathrm{FF}\circ\mathrm{Attention}$ with residual connections yields the full Transformer~\cite{vaswani2017attention}. At generation time, the next-token distribution is controlled by temperature $T$~\cite{hinton2015distilling}: $p_i = \exp(z_i/T)/\sum_j \exp(z_j/T)$, interpolating between deterministic ($T{\to}0$) and uniform ($T{\to}\infty$) sampling. Although $T$ is not a thermodynamic temperature in the strict physical sense, its structural role in the softmax mirrors that of temperature in the Boltzmann distribution, and we treat this parallel as a formal correspondence.

\section{Generative Criticality}

We apply tools from statistical field theory and intrinsic dimension estimation to probe the critical structure of LLM outputs.

\paragraph{Critical Behavior.}
We define the susceptibility of generated text as,
\begin{equation}
\chi = \frac{1}{L}\sum_{\sigma,\tau}\!\Bigl[\frac{1}{N_s}\sum_i t^{(i)}_\sigma \cdot t^{(i)}_\tau - \frac{1}{N_s^2}\sum_i t^{(i)}_\sigma\cdot \sum_j t^{(j)}_\tau\Bigr],
\end{equation}
where $L$ is the sequence length, Greek indices $\sigma, \tau$ label lattice sites (token positions), Latin indices $i, j$ label members of an ensemble of size $N_s$, and the dot products are taken in the $N$-dimensional embedding space. Although this expression is formally analogous to a variance, the non-trivial content lies in the diverging fluctuations near a critical temperature $T_c$, where $\chi \sim |T - T_c|^{-\gamma}$~\cite{kadanoff1966spin}, consistent with standard thermodynamic scaling. Similarly, the ensemble-averaged token expectation,
\begin{equation}
\langle t \rangle = \frac{1}{N_s L}\sum_{i,\sigma} t^{(i)}_\sigma
\end{equation}
serves as an order parameter~\cite{zinn2021quantum}: in the high-temperature (disordered) phase it tends toward zero; in the low-temperature (ordered) phase, deterministic generation yields a nonzero ensemble average, analogous to the spontaneous breaking of O($N$) symmetry below $T_c$.

\paragraph{Intrinsic Dimension.}
The TwoNN method estimates the intrinsic dimension $I_d$ of a data manifold from local distance ratios~\cite{facco2017estimating}. For each point, defining $\mu = r_2/r_1$, local uniformity implies $f(\mu) = I_d\,\mu^{-I_d-1}$, yielding
\begin{equation}
-\ln\bigl(1 - F(\mu)\bigr) = I_d \cdot \ln\mu\,,
\end{equation}
where $F(\mu)$ is the cumulative distribution function. The intrinsic dimension acts as an unsupervised phase-transition detector~\cite{Mendes_Santos_2021}: $I_d$ is small in the ordered phase, approaches the total degrees of freedom in the disordered phase, and exhibits non-monotonic features at criticality.

\section{Experiments and Main Results}

We use the Qwen3 family~\cite{yang2025qwen3} (0.6B--32B), fixing output length to 300 tokens. Prompts are drawn from English Wikipedia; controls include Chinese Wikipedia, jokes, poems, novels, and nonsensical texts. All generations use ``no-think'' mode (i.e., extended reasoning is disabled). For each temperature we generate $N_s = 1{,}000$ samples and compute the ensemble average over token positions to obtain the order parameter. It should be noted that the results we presented are embeddings in different dimensions(Table~\ref{Tab:1}).

\begin{figure}[htb]
  \begin{subfigure}{0.45\textwidth}
        \centering
        \includegraphics[width=\textwidth]{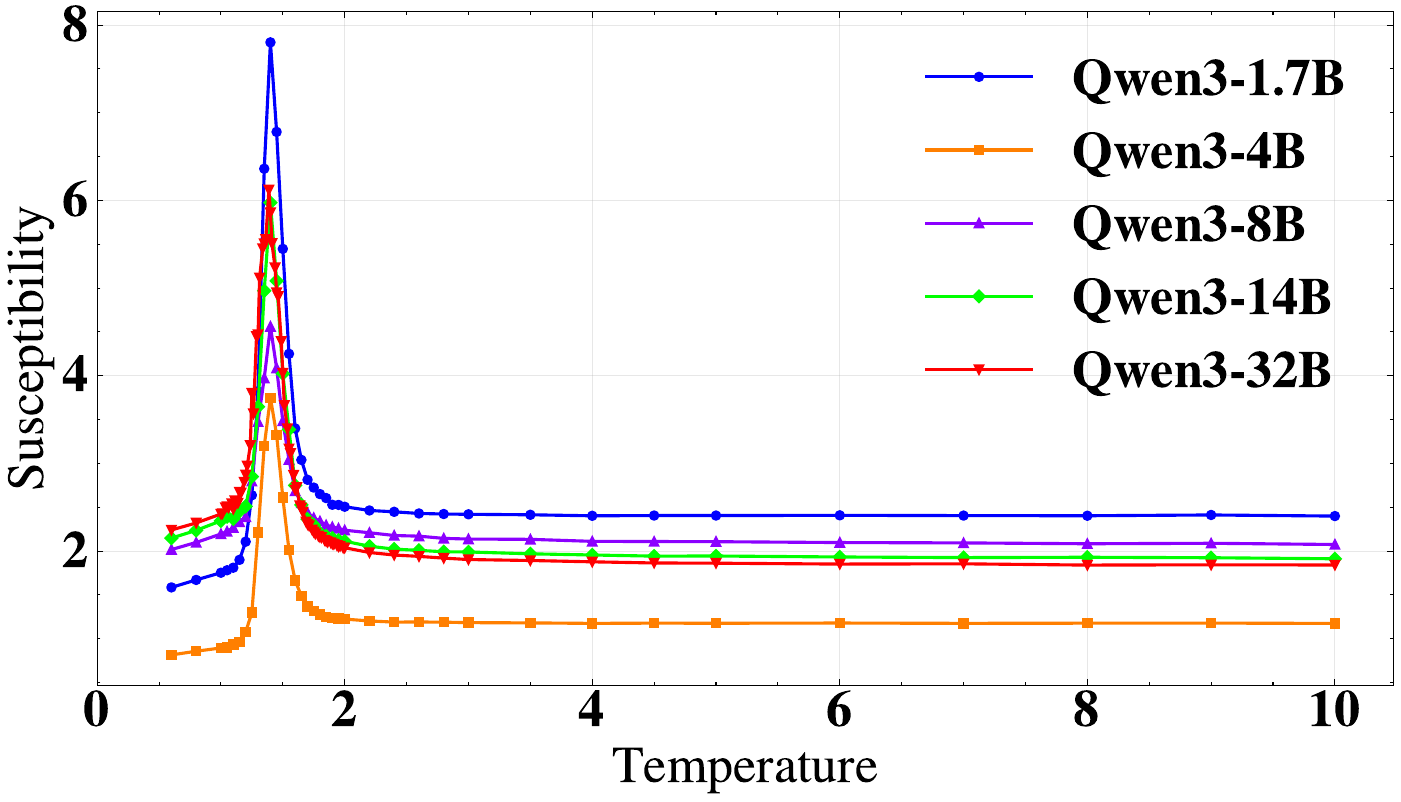}
        \caption{}
    \end{subfigure}
    \hfill
    \begin{subfigure}{0.45\textwidth}
        \centering
        \includegraphics[width=\textwidth]{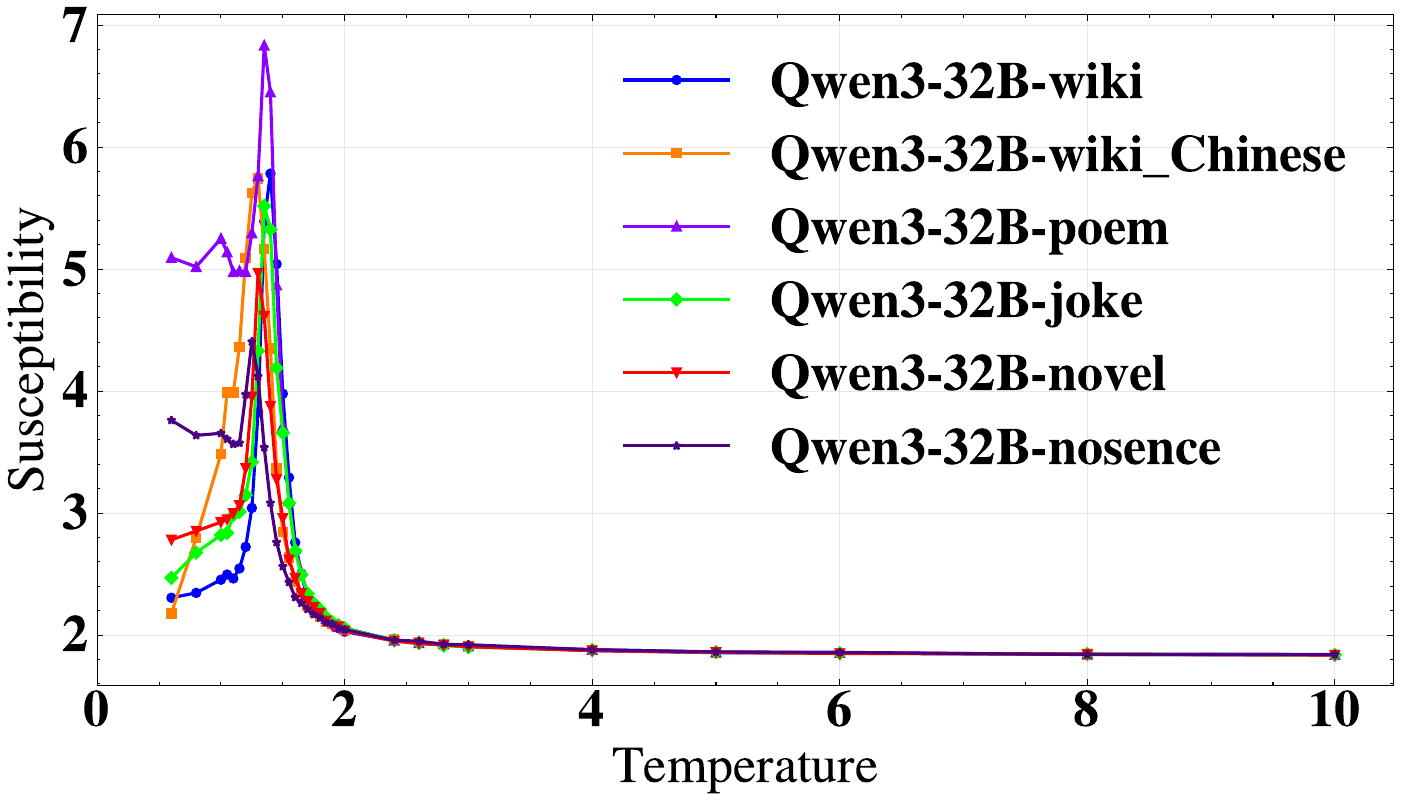}
        \caption{}
    \end{subfigure}
  \caption{Susceptibility $\chi$ vs.\ temperature $T$ for (a) different model scales using Wikipedia prompts (1000 samples, 280 tokens). All models exhibit a peak near $T_c \approx 1.4$, with larger models showing higher peak susceptibility. (b) Different prompt categories generated by Qwen3-32B (1000 samples, 300 tokens).}
  \label{fig:1}
\end{figure}

\begin{figure}[htbp]
\centering
    \begin{subfigure}{0.45\textwidth}
        \centering
        \includegraphics[width=\textwidth]{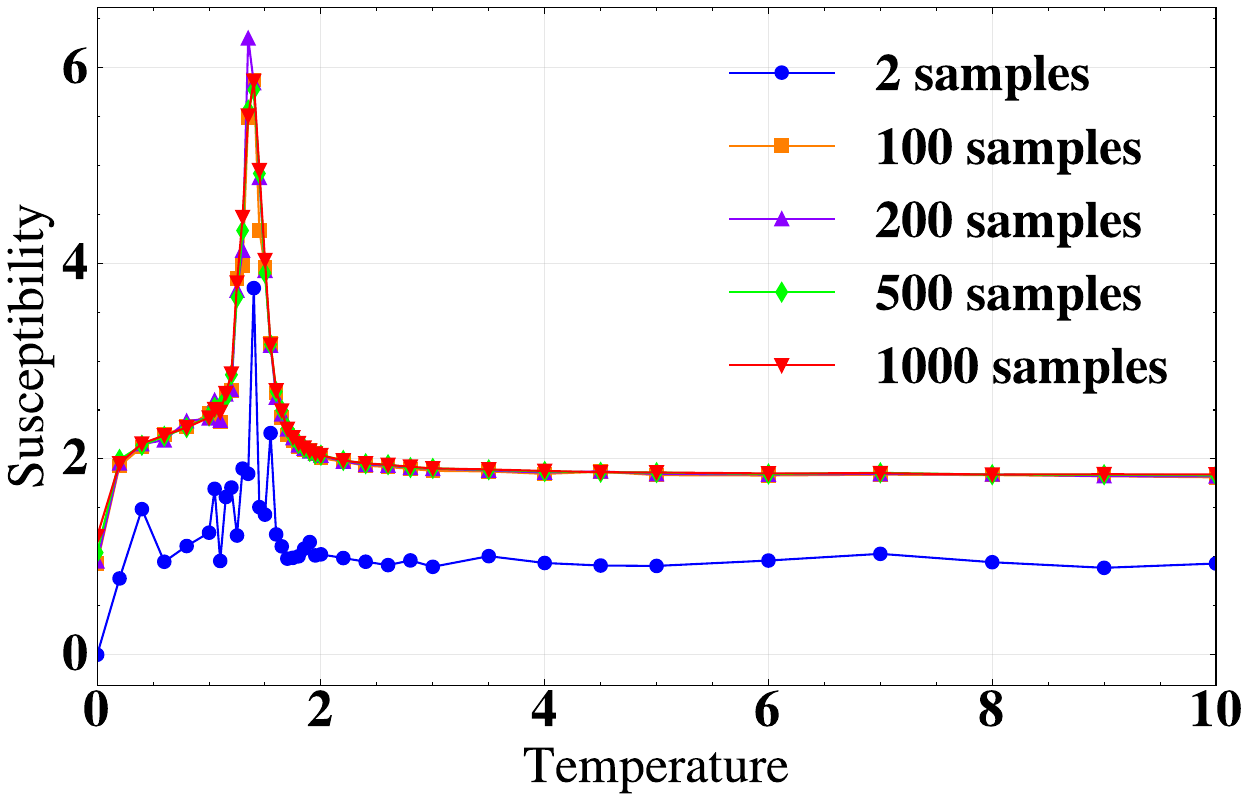}
        \caption{}
    \end{subfigure}
    \hfill
    \begin{subfigure}{0.45\textwidth}
        \centering
        \includegraphics[width=\textwidth]{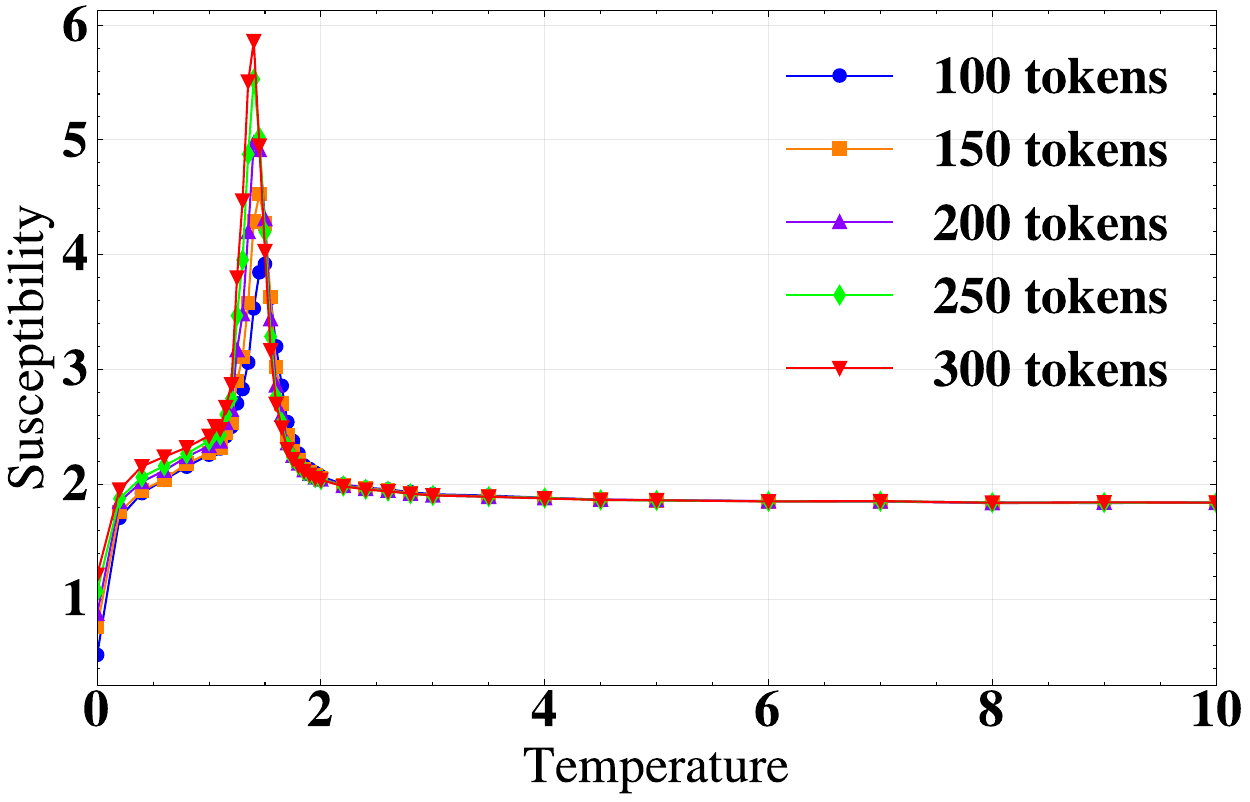}
        \caption{}
    \end{subfigure}
\caption{Stability analysis of the susceptibility signal. (a) Convergence with increasing ensemble size $N_s$ from 2 to 1000 samples; (b) Convergence with increasing sequence length from 100 to 300 tokens.}
\label{fig:2}
\end{figure}

\paragraph{Susceptibility.}
Figure~\ref{fig:1}(a) shows $\chi(T)$ for different model scales; Figure~\ref{fig:1}(b) shows results across prompt categories. A pronounced peak appears near $T_c$, with power-law scaling $\chi \sim (T-T_c)^{-\gamma}$ and $\gamma \approx 0.1$ on both sides (Figure~\ref{fig:3}. We also present the fitting results for different sizes in the appendix(Figure~\ref{fig:7}). The curves do not collapse across models or prompts, indicating that parameter count and prompt type act as distinct effective fields. Convergence with increasing ensemble size and sequence length is confirmed in Figure~\ref{fig:2}.

\begin{figure}[htbp]
    \centering
    \begin{subfigure}{0.3\textwidth}
        \centering
        \includegraphics[width=\textwidth]{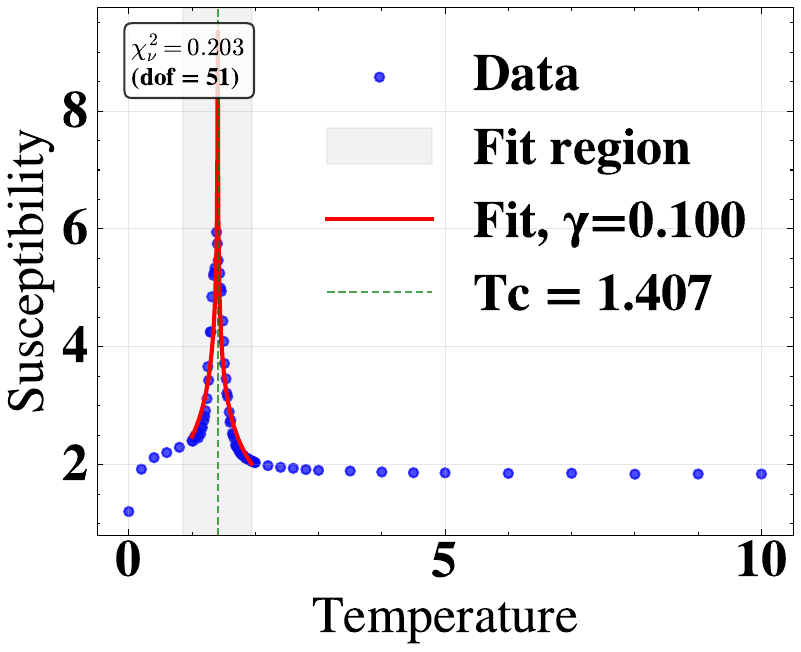}
        \caption{}
    \end{subfigure}
    \hfill
    \begin{subfigure}{0.3\textwidth}
        \centering
        \includegraphics[width=\textwidth]{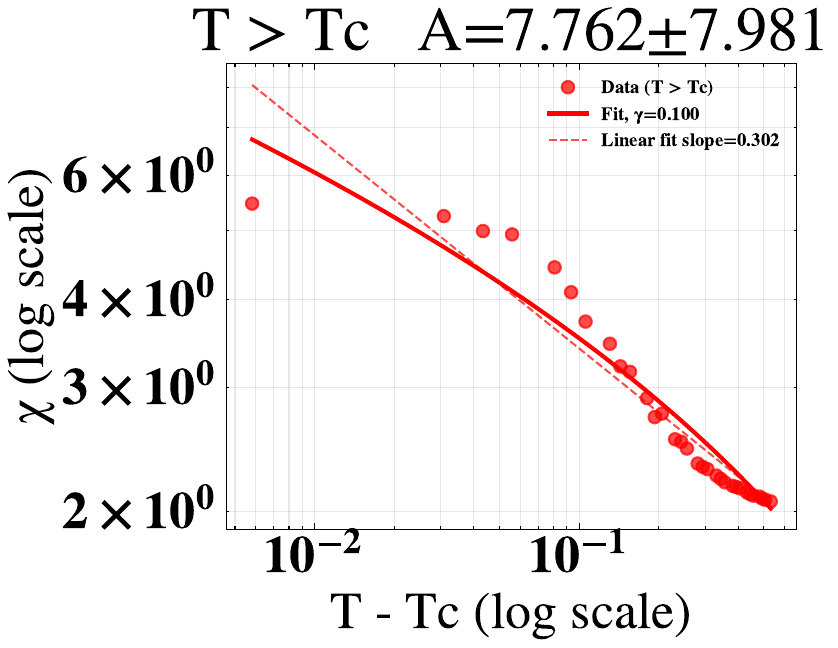}
        \caption{}
    \end{subfigure}
    \hfill
    \begin{subfigure}{0.3\textwidth}
        \centering
        \includegraphics[width=\textwidth]{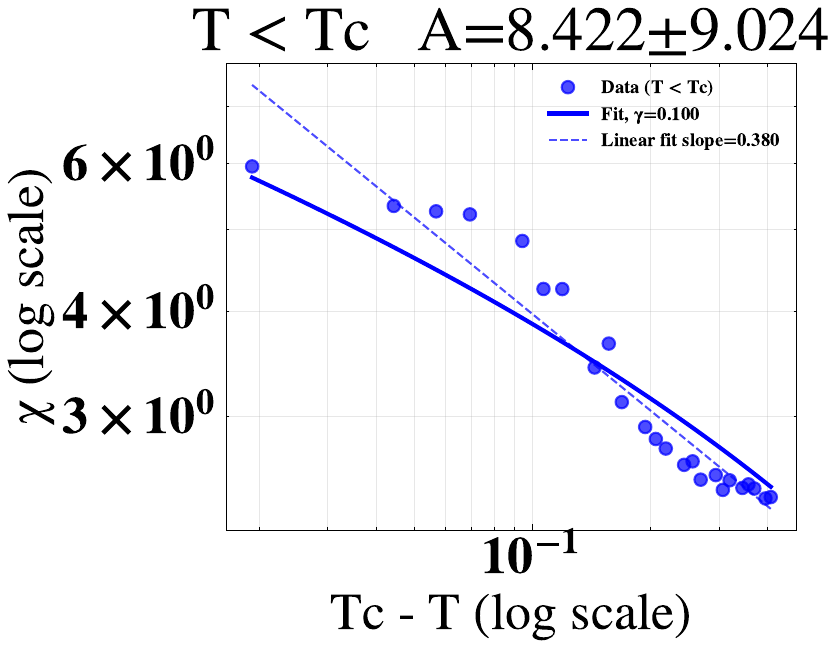}
        \caption{}
    \end{subfigure}
    \caption{Power-law fits of susceptibility near $T_c$ for Qwen3-32B. (a) Full $\chi(T)$ curve with the fit region highlighted; (b) log-log plot for $T > T_c$; (c) log-log plot for $T < T_c$. Both sides yield a critical exponent $\gamma \approx 0.1$, consistent with power-law divergence. }
    \label{fig:3}
\end{figure}

\begin{figure}[htbp]
\centering
    \begin{subfigure}{0.4\textwidth}
        \centering
        \includegraphics[width=\textwidth]{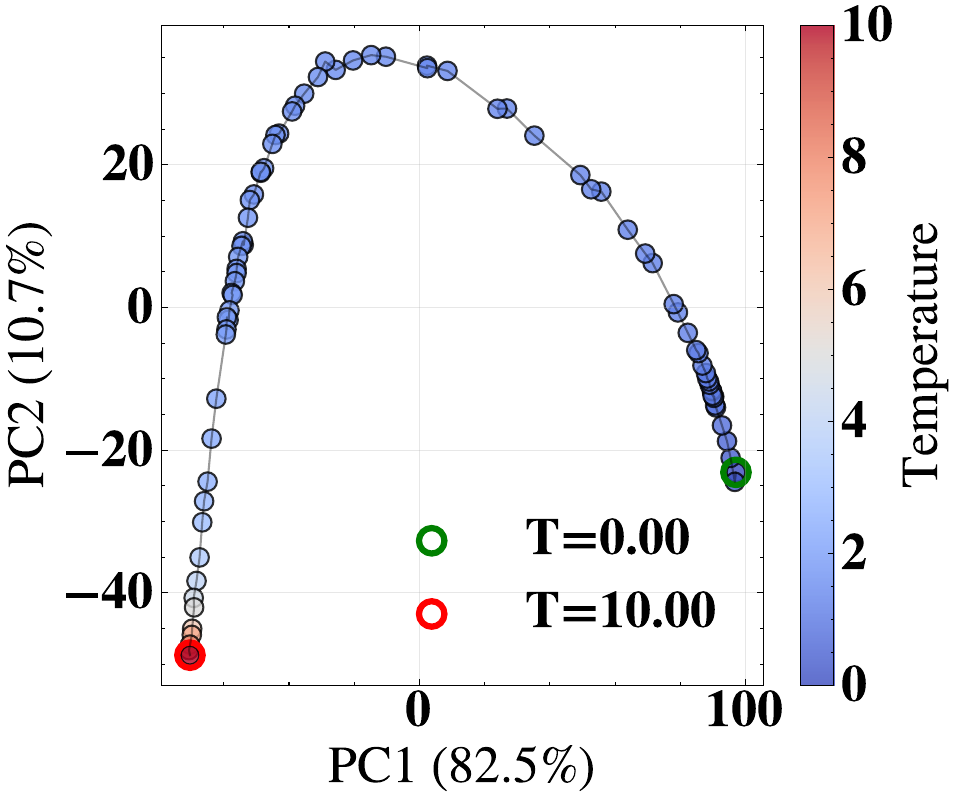}
        \caption{}
    \end{subfigure}
    \hfill
    \begin{subfigure}{0.5\textwidth}
        \centering
        \includegraphics[width=\textwidth]{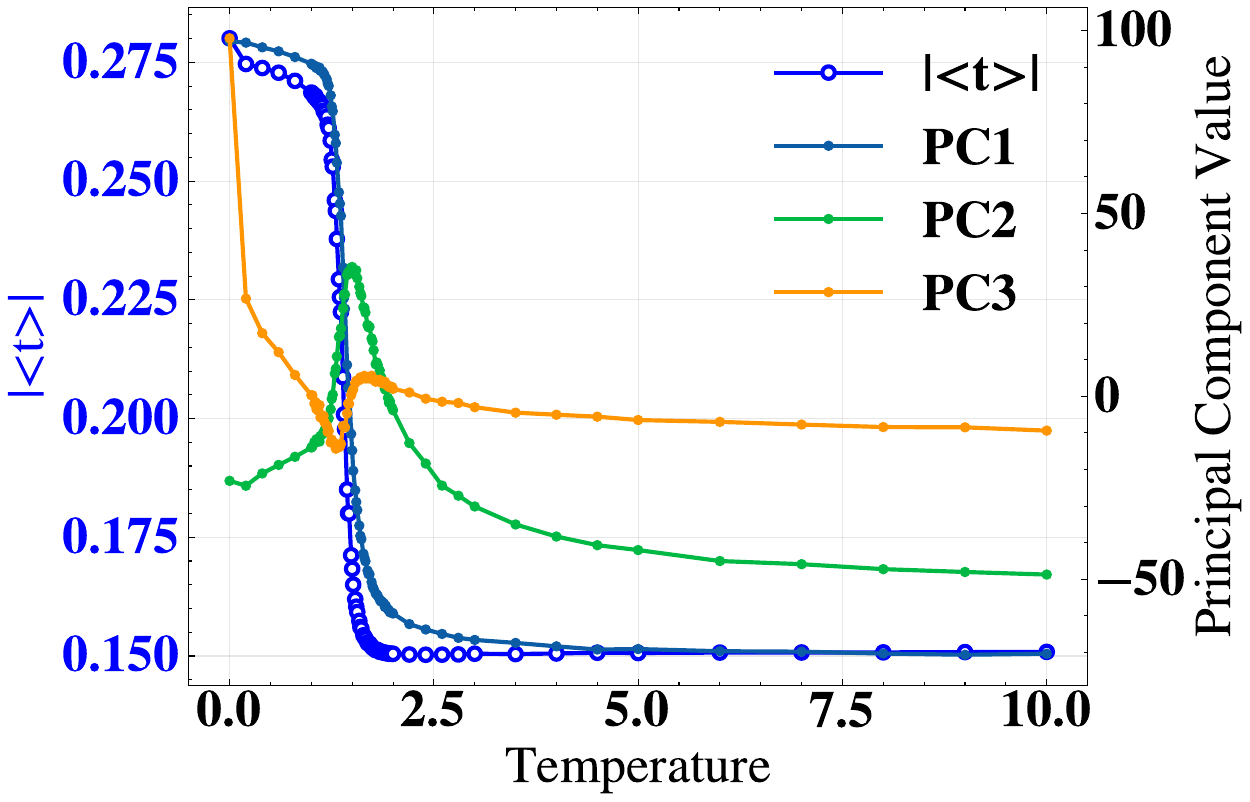}
        \caption{}
    \end{subfigure}
\caption{Order parameter structure via PCA. (a) Projection onto the first two principal components (PC1 explains 82.5\%, PC2 explains 10.7\%), colored by temperature. Below $T_c$, ensemble outputs cluster tightly along a single direction; above $T_c$, they disperse. (b) Magnitude $|\langle t \rangle|$ and the first three principal component values vs.\ temperature, showing a sharp transition near $T_c$.}
\label{fig:4}
\end{figure}

\paragraph{Order Parameter Structure.}
Applying PCA to the ensemble-averaged order-parameter vector across temperatures, we find that generated texts concentrate along a single semantic direction below $T_c$, with a sharp directional change at criticality (Figure~\ref{fig:4}). This confirms the phase transition and suggests that, at low temperatures, the LLM effectively selects tokens approximating a unique semantic target.

\begin{figure}[htbp]
    \centering
    \includegraphics[width=0.6\textwidth]{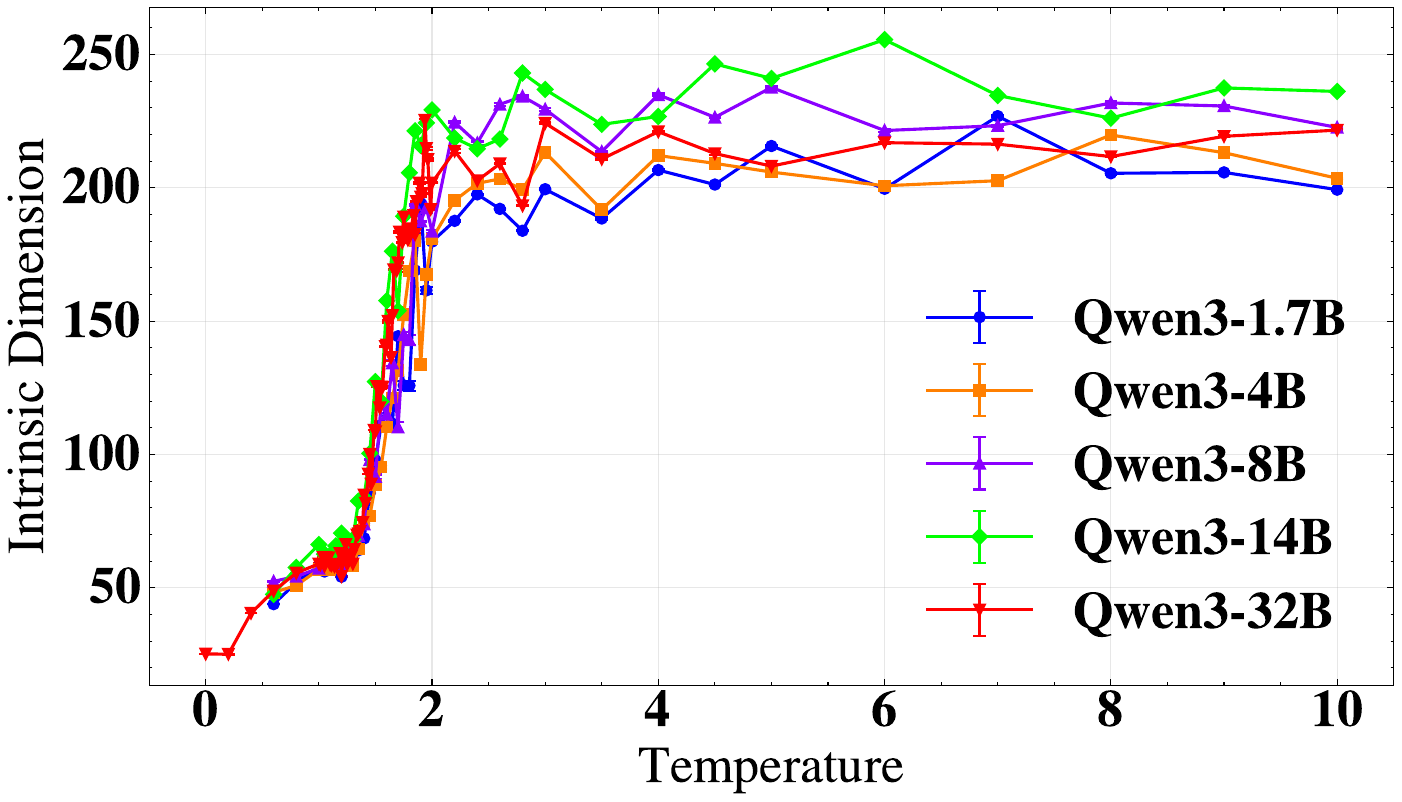}
    \caption{Intrinsic dimension $I_d$ vs.\ temperature for five Qwen3 model scales. All models exhibit a minimum near $T_c \approx 1.3$--$1.5$, after which $I_d$ rises sharply and saturates at high $T$, consistent with the onset of a disordered phase where the configuration space expands.}
    \label{fig:5}
\end{figure}

\paragraph{Intrinsic Dimension.}
We apply the TwoNN method to sentence-level embedding vectors (obtained via average pooling over token embeddings) of $N_s = 1{,}000$ texts at each temperature (Figure~\ref{fig:5}). The intrinsic dimension $I_d$ reaches a minimum near $T_c$ and rises rapidly above it: at low $T$ the configuration space is restricted; near $T_c$ it enters a critical regime; at high $T$ the system explores the full configuration space. Fits are excellent at all nonzero temperatures (see Appendix~A).

\section{Conclusion}
We have presented a statistical-field framework for LLM-generated text, defining susceptibility and an order parameter in the token embedding space. Systematic variation of the softmax temperature reveals phase-transition-like behavior: a divergent susceptibility peak near $T_c$, a sharp change in the order parameter accompanied by dimensional collapse onto a single semantic direction, and a minimum in the TwoNN intrinsic dimension in the same critical region. These mutually consistent signatures validate the feasibility of statistical-field analysis for probing the collective structure of LLM outputs. Future directions include applying renormalization-group methods~\cite{wilson1974renormalization} to capture long-range correlations, and leveraging the critical behavior identified here as a diagnostic for trustworthy AI—for instance, the sharp onset of the disordered phase may serve as a quantitative indicator of regimes in which LLM outputs become unreliable.

\section*{Acknowledgement}
We thank Drs. Kai Nakaishi, Jan Pawlowski, Enrico Rinaldi, Gabriele Di Ubaldo, and Sho Yokoi for helpful discussions.
We thank the DEEP-IN working group at RIKEN-iTHEMS for support in the preparation of this paper.
XG and HR is supported by the National Natural Science Foundation of China under Grant No. 12035007.
LX is supported by JSPS KAKENHI Grant No. 25H01560, and JST-BOOST Grant No.JPMJBY24H9.
 
\bibliographystyle{unsrt}  
\bibliography{ref}

@article{Mendes_Santos_2021,
   title={Unsupervised Learning Universal Critical Behavior via the Intrinsic Dimension},
   volume={11},
   ISSN={2160-3308},
   url={http://dx.doi.org/10.1103/PhysRevX.11.011040},
   DOI={10.1103/physrevx.11.011040},
   number={1},
   journal={Physical Review X},
   publisher={American Physical Society (APS)},
   author={Mendes-Santos, T. and Turkeshi, X. and Dalmonte, M. and Rodriguez, Alex},
   year={2021},
   month=feb }

@article{sun2025phase,
  title={Phase Transitions in Large Language Models and the $ O (N) $ Model},
  author={Sun, Youran and Haghighat, Babak},
  journal={arXiv preprint arXiv:2501.16241},
  year={2025}
}

@article{nakaishi2024critical,
  title={Critical phase transition in large language models},
  author={Nakaishi, Kai and Nishikawa, Yoshihiko and Hukushima, Koji},
  journal={arXiv preprint arXiv:2406.05335},
  year={2024}
}

@book{amit2005field,
  title={Field theory, the renormalization group, and critical phenomena: graphs to computers},
  author={Amit, Daniel J and Martin-Mayor, Victor},
  year={2005},
  publisher={World Scientific}
}

@article{vaswani2017attention,
  title={Attention is all you need},
  author={Vaswani, Ashish and Shazeer, Noam and Parmar, Niki and Uszkoreit, Jakob and Jones, Llion and Gomez, Aidan N and Kaiser, {\L}ukasz and Polosukhin, Illia},
  journal={Advances in neural information processing systems},
  volume={30},
  year={2017}
}

@book{bishop2023deep,
  title={Deep learning: Foundations and concepts},
  author={Bishop, Christopher M and Bishop, Hugh},
  year={2023},
  publisher={Springer Nature}
}

@article{yang2023teal,
  title={Teal: Tokenize and embed all for multi-modal large language models},
  author={Yang, Zhen and Zhang, Yingxue and Meng, Fandong and Zhou, Jie},
  journal={arXiv preprint arXiv:2311.04589},
  year={2023}
}

@article{mehta2023semantic,
  title={Semantic tokenizer for enhanced natural language processing},
  author={Mehta, Sandeep and Shah, Darpan and Kulkarni, Ravindra and Caragea, Cornelia},
  journal={arXiv preprint arXiv:2304.12404},
  year={2023}
}

@incollection{egger2022text,
  title={Text representations and word embeddings: Vectorizing textual data},
  author={Egger, Roman},
  booktitle={Applied data science in tourism: Interdisciplinary approaches, methodologies, and applications},
  pages={335--361},
  year={2022},
  publisher={Springer}
}

@article{hinton2015distilling,
  title={Distilling the knowledge in a neural network},
  author={Hinton, Geoffrey and Vinyals, Oriol and Dean, Jeff},
  journal={arXiv preprint arXiv:1503.02531},
  year={2015}
}

@article{coleman1974spontaneous,
  title={Spontaneous symmetry breaking in the O (N) model for large N},
  author={Coleman, Sidney and Jackiw, Roman and Politzer, HDavid},
  journal={Physical Review D},
  volume={10},
  number={8},
  pages={2491},
  year={1974},
  publisher={APS}
}

@article{eynard1995exact,
  title={Exact solution of the O (n) model on a random lattice},
  author={Eynard, Bertrand and Kristjansen, Charlotte},
  journal={Nuclear Physics B},
  volume={455},
  number={3},
  pages={577--618},
  year={1995},
  publisher={Elsevier}
}

@article{yang2025qwen3,
  title={Qwen3 technical report},
  author={Yang, An and Li, Anfeng and Yang, Baosong and Zhang, Beichen and Hui, Binyuan and Zheng, Bo and Yu, Bowen and Gao, Chang and Huang, Chengen and Lv, Chenxu and others},
  journal={arXiv preprint arXiv:2505.09388},
  year={2025}
}

@book{zinn2021quantum,
  title={Quantum field theory and critical phenomena},
  author={Zinn-Justin, Jean},
  volume={171},
  year={2021},
  publisher={Oxford university press}
}

@article{chang2023simple,
  title={A simple explanation for the phase transition in large language models with list decoding},
  author={Chang, Cheng-Shang},
  journal={arXiv preprint arXiv:2303.13112},
  year={2023}
}

@article{Scheibner2025LargeLM,
  title={Large language models and the entropy of English},
  author={Colin Scheibner and Lindsay M. Smith and William Bialek},
  journal={ArXiv},
  year={2025},
  volume={abs/2512.24969},
  url={https://api.semanticscholar.org/CorpusID:284351303}
}

@inproceedings{zhu2024hot,
  title={Hot or cold? adaptive temperature sampling for code generation with large language models},
  author={Zhu, Yuqi and Li, Jia and Li, Ge and Zhao, YunFei and Jin, Zhi and Mei, Hong},
  booktitle={Proceedings of the AAAI Conference on Artificial Intelligence},
  volume={38},
  number={1},
  pages={437--445},
  year={2024}
}

@article{carleo2019machine,
  title={Machine learning and the physical sciences},
  author={Carleo, Giuseppe and Cirac, Ignacio and Cranmer, Kyle and Daudet, Laurent and Schuld, Maria and Tishby, Naftali and Vogt-Maranto, Leslie and Zdeborov{\'a}, Lenka},
  journal={Reviews of Modern Physics},
  volume={91},
  number={4},
  pages={045002},
  year={2019},
  publisher={APS}
}

@article{wilson1974renormalization,
  title={The renormalization group and the $\epsilon$ expansion},
  author={Wilson, Kenneth G and Kogut, John},
  journal={Physics Reports},
  volume={12},
  number={2},
  pages={75--199},
  year={1974},
  publisher={Elsevier}
}

@article{kadanoff1966spin,
  title={Spin-spin correlations in the two-dimensional ising model},
  author={Kadanoff, Leo P},
  journal={Il Nuovo Cimento B (1965-1970)},
  volume={44},
  number={2},
  pages={276--305},
  year={1966},
  publisher={Springer}
}

@article{li2018measuring,
  title={Measuring the intrinsic dimension of objective landscapes},
  author={Li, Chunyuan and Farkhoor, Heerad and Liu, Rosanne and Yosinski, Jason},
  journal={arXiv preprint arXiv:1804.08838},
  year={2018}
}

@article{takahira2016entropy,
  title={Entropy rate estimates for natural language—a new extrapolation of compressed large-scale corpora},
  author={Takahira, Ryosuke and Tanaka-Ishii, Kumiko and D{\k{e}}bowski, {\L}ukasz},
  journal={Entropy},
  volume={18},
  number={10},
  pages={364},
  year={2016},
  publisher={MDPI}
}

@inproceedings{li1989mutual,
  title={Mutual information functions of natural language texts},
  author={Li, Wentian},
  year={1989},
  organization={Santa Fe Institute Santa Fe, NM, USA}
}

@article{lin2017critical,
  title={Critical behavior in physics and probabilistic formal languages},
  author={Lin, Henry W and Tegmark, Max},
  journal={Entropy},
  volume={19},
  number={7},
  pages={299},
  year={2017},
  publisher={MDPI}
}

@book{pinker2003language,
  title={The language instinct: How the mind creates language},
  author={Pinker, Steven},
  year={2003},
  publisher={Penguin uK}
}

@book{chomsky2002syntactic,
  title={Syntactic structures},
  author={Chomsky, Noam},
  year={2002},
  publisher={Walter de Gruyter}
}

@article{mitchell2023debate,
  title={The debate over understanding in AI’s large language models},
  author={Mitchell, Melanie and Krakauer, David C},
  journal={Proceedings of the National Academy of Sciences},
  volume={120},
  number={13},
  pages={e2215907120},
  year={2023},
  publisher={National Academy of Sciences}
}

@article{facco2017estimating,
  title={Estimating the intrinsic dimension of datasets by a minimal neighborhood information},
  author={Facco, Elena and d’Errico, Maria and Rodriguez, Alex and Laio, Alessandro},
  journal={Scientific reports},
  volume={7},
  number={1},
  pages={12140},
  year={2017},
  publisher={Nature Publishing Group UK London}
}

\newpage
\appendix
\section{TwoNN Fitting}

We present the data and fitting plot for the TwoNN method applied to Qwen3‑32B in the appendix to show more details.

\begin{figure}[htbp]
    \centering
    \includegraphics[width=0.9\textwidth]{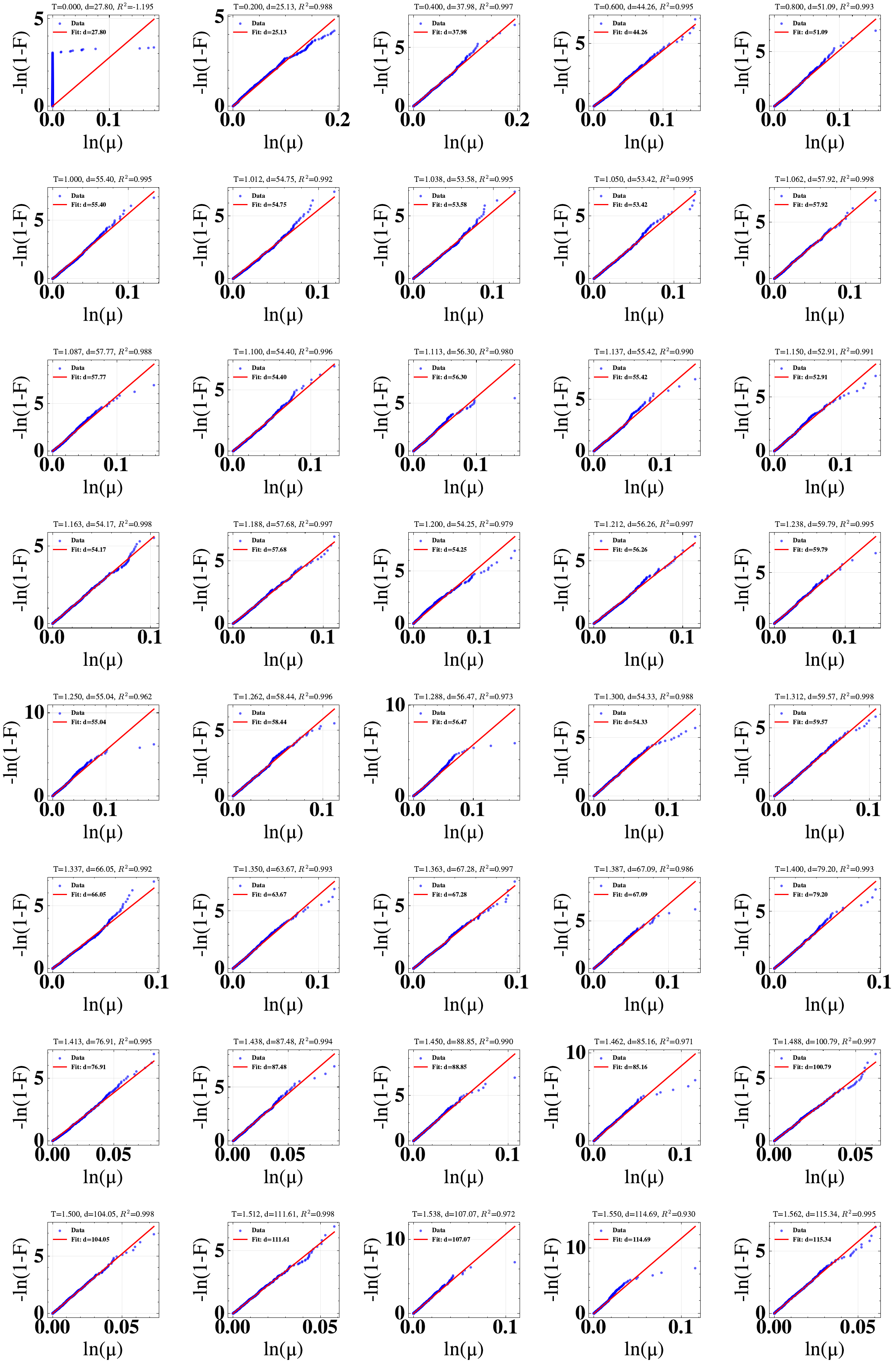}
    \label{fig:6}
    \
\end{figure}
\begin{figure}[htbp]
    \centering
    \includegraphics[width=0.9\textwidth]{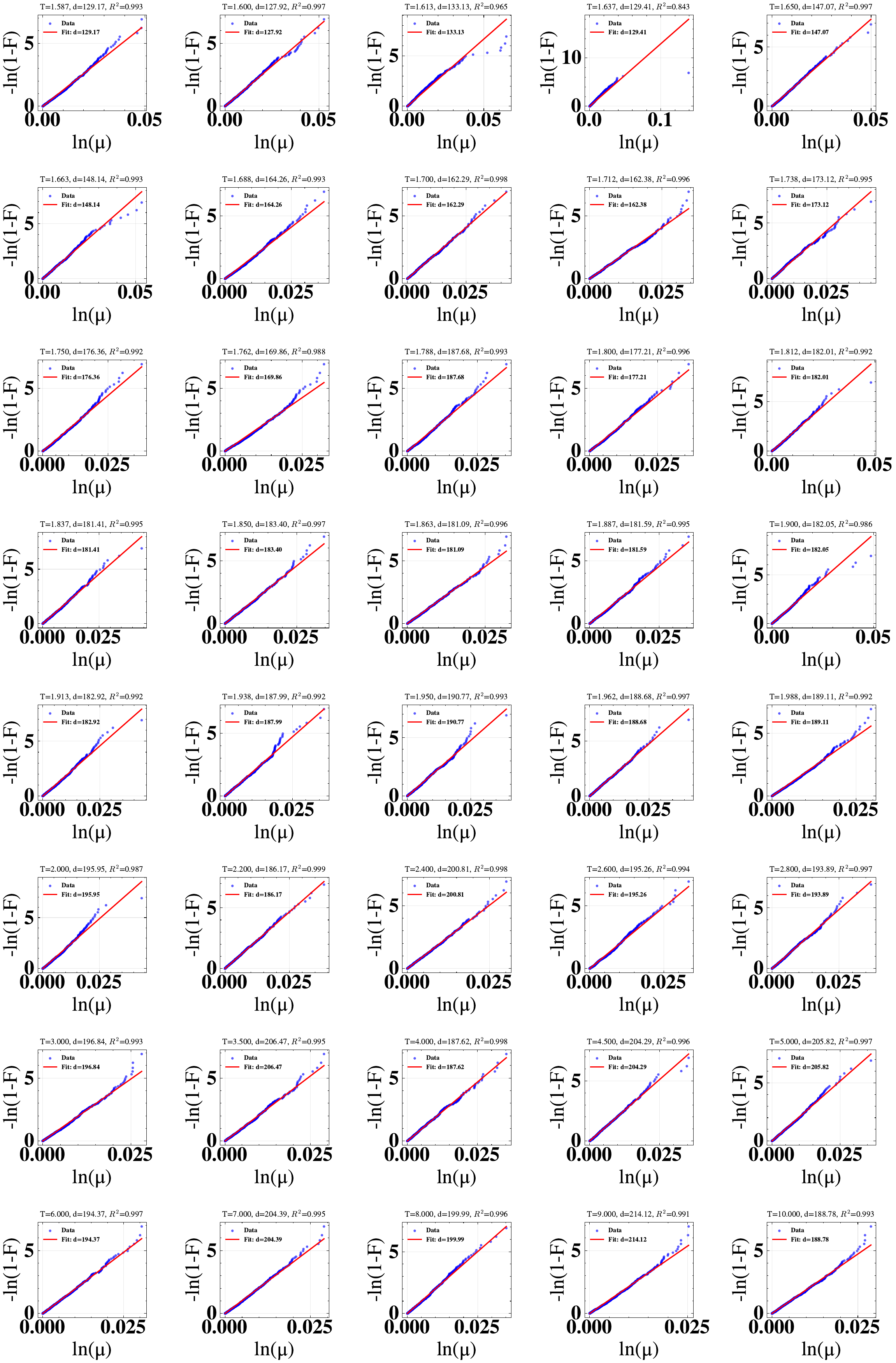}
    \caption{By evaluating the ratio of the second-nearest neighbor distance to the nearest neighbor distance among these data points and fitting its distribution, the derived slope yields the intrinsic dimension.}
    \label{fig:7}
\end{figure}

\newpage
\section{Configurations of Qwen3 models}
We include a table to illustrate their varying semantic spaces across different embedding dimensions.
\begin{table}[htbp]
\centering
\begin{tabular}{lcccccc}
\toprule
\textbf{Model} &  \textbf{Hidden size} & \textbf{Layers} & \textbf{Attention heads (Q / KV)} \\
\midrule
Qwen3-0.6B  & 1024 & 28 & 16 / 8 \\
Qwen3-1.7B  & 2048 & 28 & 16 / 8 \\
Qwen3-4B    & 2560 & 36 & 32 / 8 \\
Qwen3-8B    & 4096 & 36 & 32 / 8 \\
Qwen3-14B   & 5120 & 48 & 40 / 8 \\
Qwen3-32B   & 8192 & 64 & 64 / 8 \\
\bottomrule
\end{tabular}
\caption{Architectural parameters of the Qwen3 series models.}
\label{Tab:1}
\end{table}

\section{Testing scaling laws at different sizes}

\begin{figure}[htbp]
    \centering
    \begin{subfigure}{0.3\textwidth}
        \centering
        \includegraphics[width=\textwidth]{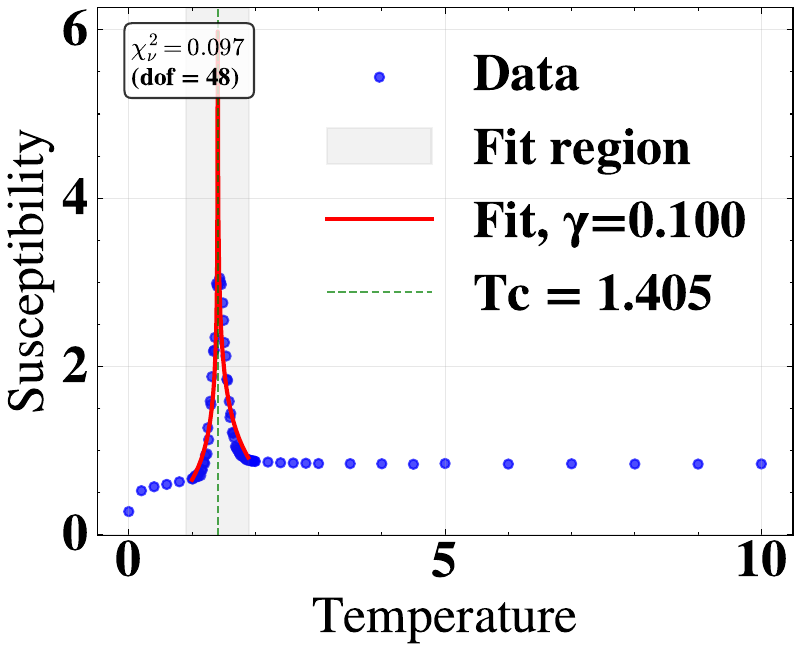}
        \caption{}
    \end{subfigure}
    \hfill
    \begin{subfigure}{0.3\textwidth}
        \centering
        \includegraphics[width=\textwidth]{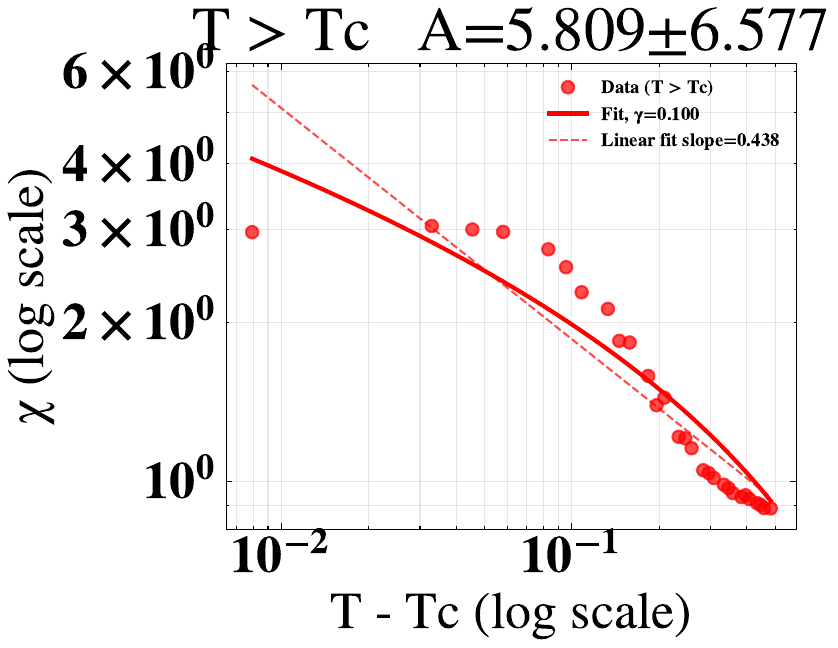}
        \caption{}
    \end{subfigure}
    \hfill
    \begin{subfigure}{0.3\textwidth}
        \centering
        \includegraphics[width=\textwidth]{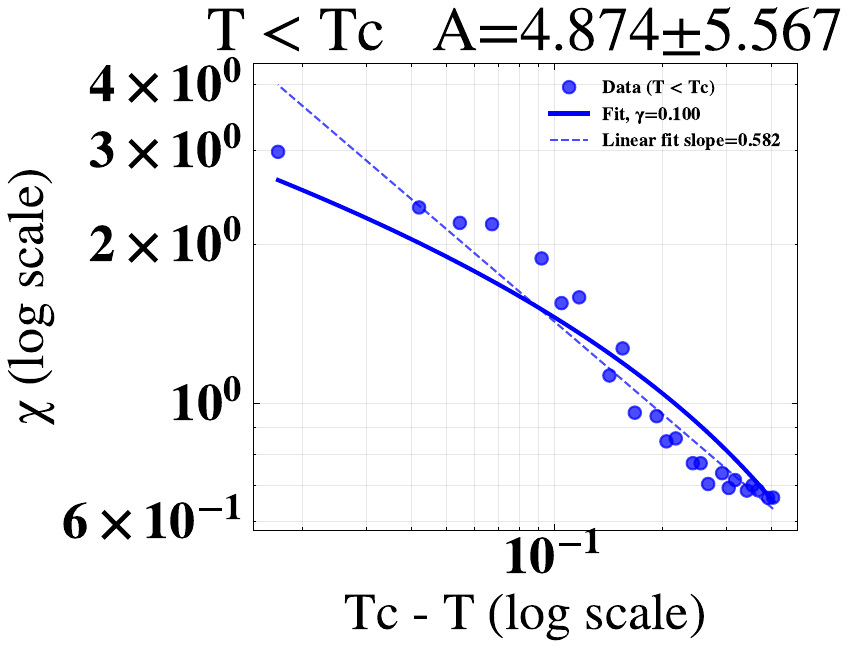}
        \caption{}
    \end{subfigure}
    \centering
    \begin{subfigure}{0.3\textwidth}
        \centering
        \includegraphics[width=\textwidth]{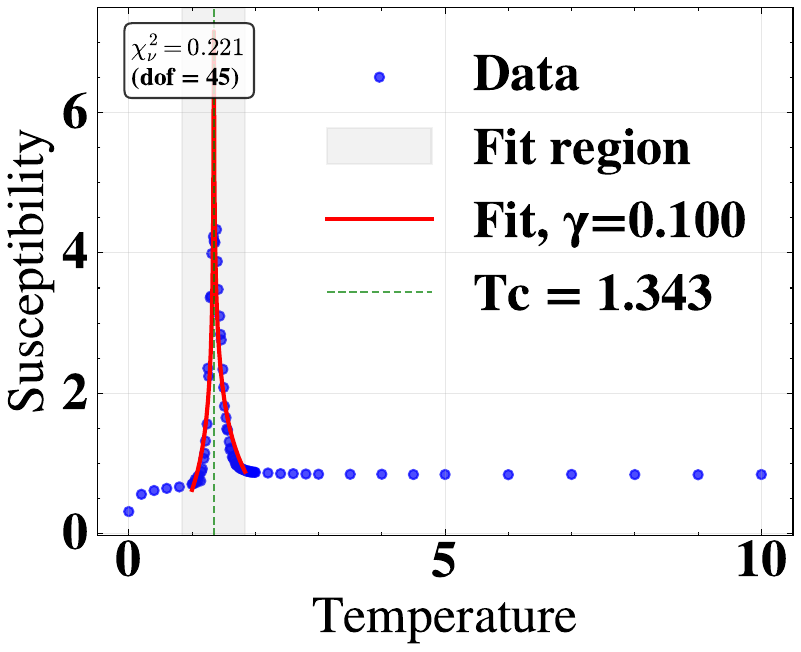}
        \caption{}
    \end{subfigure}
    \hfill
    \begin{subfigure}{0.3\textwidth}
        \centering
        \includegraphics[width=\textwidth]{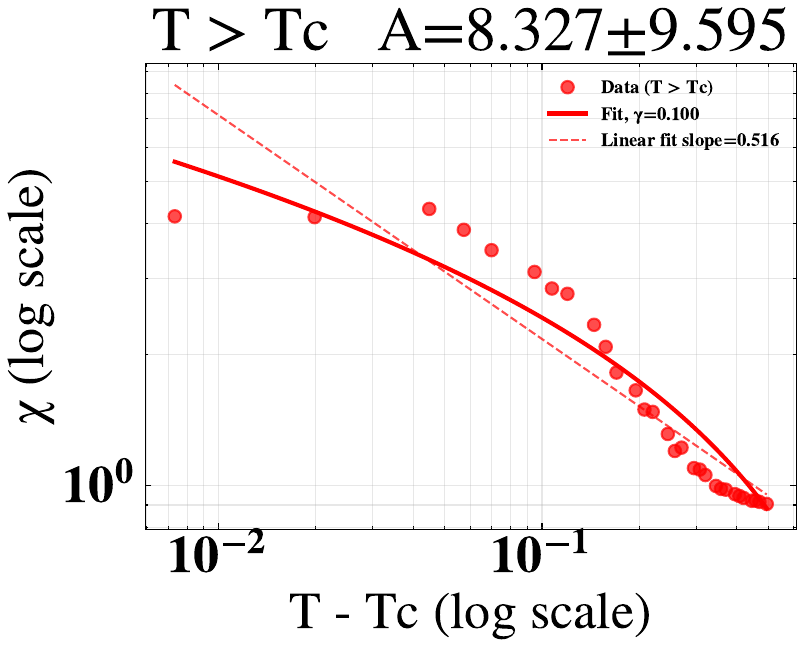}
        \caption{}
    \end{subfigure}
    \hfill
    \begin{subfigure}{0.3\textwidth}
        \centering
        \includegraphics[width=\textwidth]{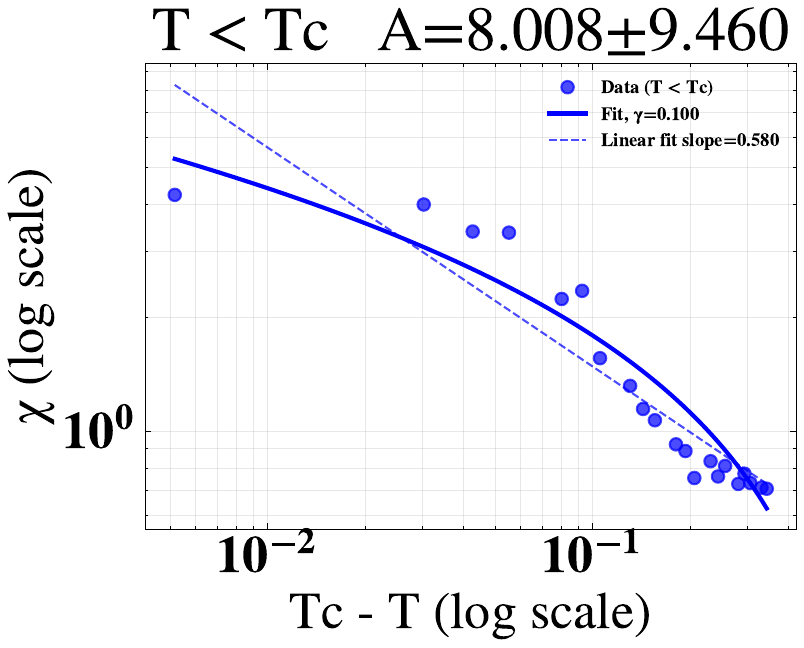}
        \caption{}
    \end{subfigure}
    \centering
    \begin{subfigure}{0.3\textwidth}
        \centering
        \includegraphics[width=\textwidth]{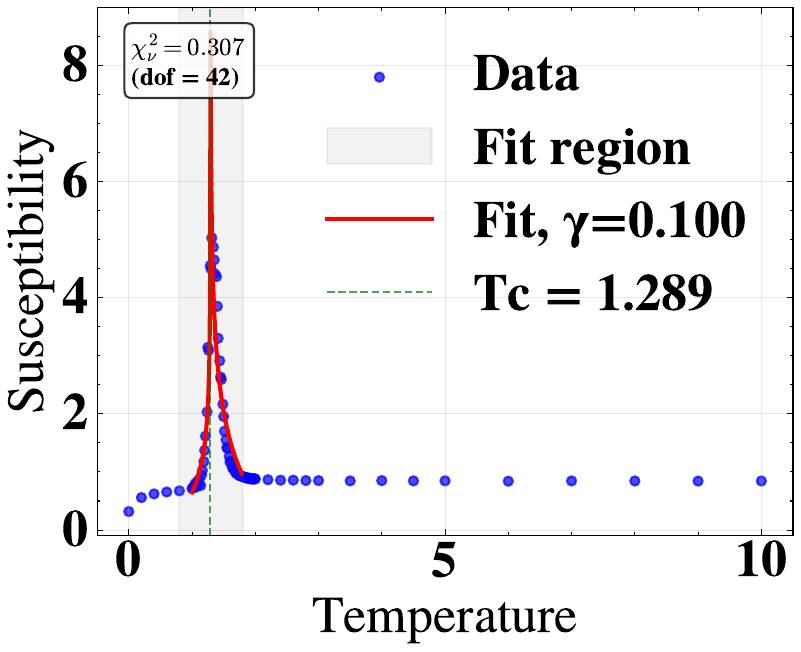}
        \caption{}
    \end{subfigure}
    \hfill
    \begin{subfigure}{0.3\textwidth}
        \centering
        \includegraphics[width=\textwidth]{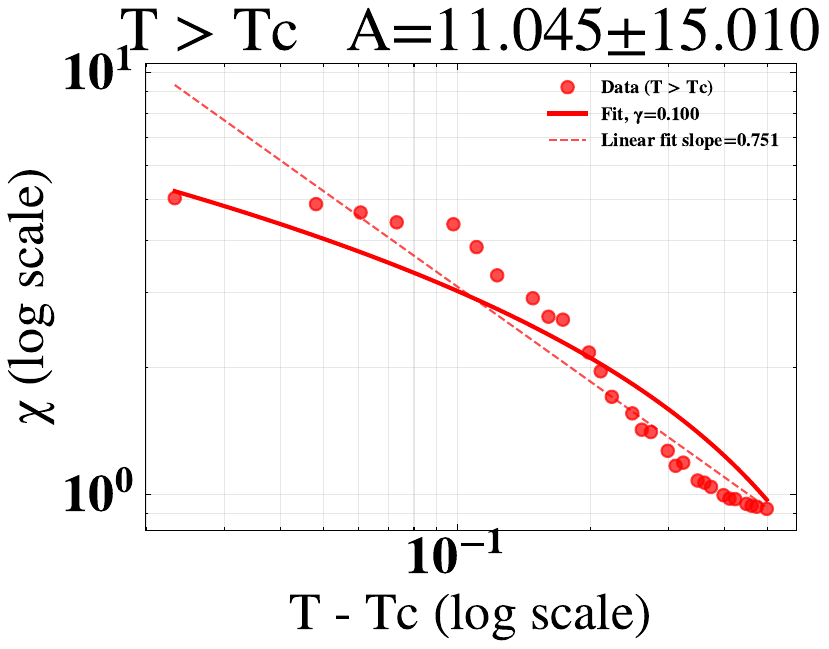}
        \caption{}
    \end{subfigure}
    \hfill
    \begin{subfigure}{0.3\textwidth}
        \centering
        \includegraphics[width=\textwidth]{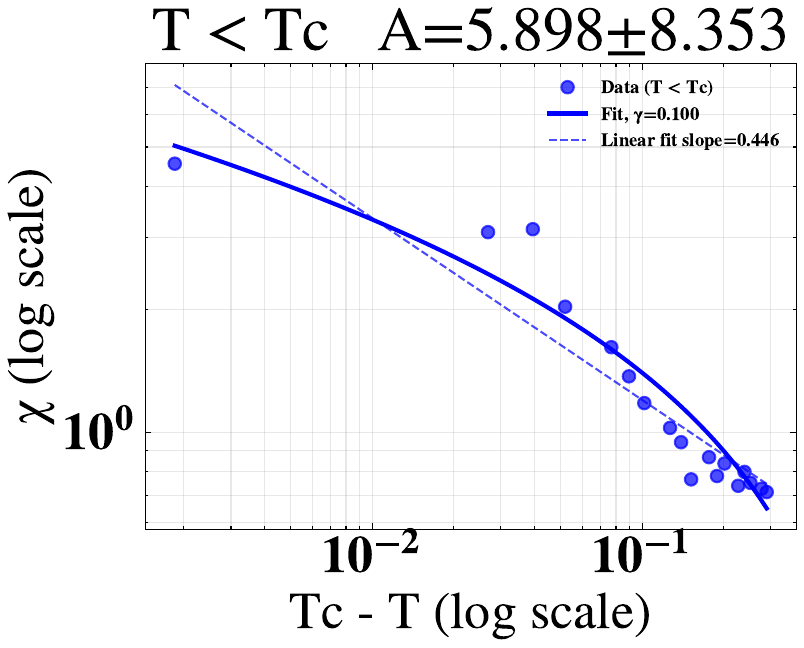}
        \caption{}
    \end{subfigure}
    \caption{Scaling laws tests at different sizes. Panels (a), (b), and (c) show exponent fitting for a system size of 200 tokens; panels (d), (e), and (f) for 400 tokens; panels (g), (h), and (i) for 500 tokens.}
    \label{fig:7}
\end{figure}



\end{document}